\documentclass[11pt]{article}

\usepackage[most]{tcolorbox}

\usepackage{times}
\usepackage[T1]{fontenc}
\usepackage[utf8]{inputenc}

\usepackage{listings}
\usepackage{xcolor}
\usepackage{verbatim}

\definecolor{jsonbg}{RGB}{248,248,248}
\definecolor{jsonstring}{RGB}{163,21,21}
\definecolor{framegray}{RGB}{220,220,220}
\lstdefinelanguage{json}{
    basicstyle=\ttfamily\small,
    showstringspaces=false,
    breaklines=true,
    frame=none,
    backgroundcolor={},
    string=[s]{"}{"},
    stringstyle=\color{jsonstring},
    comment=[l]{//},
    morecomment=[s]{/*}{*/},
    escapeinside={(*}{*)},
}

\newtcblisting{jsonbox}{
    sharp corners,
    enhanced,
    boxrule=0.5pt,
    colframe=framegray,
    colback=jsonbg,
    left=4mm, right=4mm, top=2mm, bottom=2mm,
    listing only,
    listing options={language=json}
}

\usepackage{acl}
\usepackage{latexsym}
\usepackage{microtype}
\usepackage{inconsolata}
\usepackage{graphicx}
\usepackage{amsmath}
\usepackage{booktabs}
\usepackage{multirow}
\usepackage{tabularx}
\usepackage{caption}

\newcommand{\safeincludegraphics}[2][]{%
  \IfFileExists{#2}{%
    \includegraphics[#1]{#2}%
  }{%
    \fbox{\parbox[c][0.22\textheight][c]{0.46\textwidth}{\centering Missing figure: \\ \texttt{\detokenize{#2}}}}%
  }%
}

\title{IndicGuard: A Multilingual Safety Guard Model and Dataset for Indic Languages}

\author{%
  \textbf{Parth Bramhecha,}
  \textbf{Smit Deshmukh,}
  \textbf{Sairaj Bodhale,}
  \textbf{Adwait Borate,}
  \textbf{Raviraj Joshi} \\
  \vspace{3pt}
  \textit{L3Cube-Labs, Pune} \\
  \vspace{3pt}
  \small\texttt{\{}\href{mailto:parth.bramhecha007@gmail.com}{\texttt{parth.bramhecha007}}\texttt{,}
  \href{mailto:deshmukhsmit11@gmail.com}{\texttt{deshmukhsmit11}}\texttt{,}
  \href{mailto:adwaitborate@gmail.com}{\texttt{sairaj.sab}}\texttt{,}
  \href{mailto:sairaj.sab@gmail.com}{\texttt{adwaitborate}}\texttt{,}
  \href{mailto:ravirajoshi@gmail.com}{\texttt{ravirajoshi}}\texttt{\}@gmail.com}
}

\begin{document}
\maketitle
\begin{abstract}
As Large Language Models (LLMs) achieve widespread integration across diverse linguistic landscapes, ensuring their safety and alignment with regional normative values remains a critical challenge. Current safety mechanisms are predominantly optimized for English-centric frameworks, often failing to capture the unique socio-cultural sensitivities and localized categories of harm inherent to the Indic region. To address this gap, we introduce \textbf{IndicGuard}, a multilingual safety guard model and dataset for Indic languages. We construct a high-volume, culturally nuanced safety dataset encompassing ten major Indic languages, systematically curated to capture regional harms, sensitive socio-political contexts, and adversarial jailbreaks. Leveraging this corpus, we fine-tune a 4B-parameter instruction-tuned model based on Gemma-3-4B-IT to serve as a multilingual safety guardrail for real-time content moderation and policy compliance checking. Our empirical evaluations demonstrate that IndicGuard significantly enhances LLM robustness against localized vulnerabilities, achieving high moderation consistency across different conversational turns. Crucially, IndicGuard consistently outperforms the existing baseline model, CultureGuard, across evaluated languages. Finally, we demonstrate that our model effectively generalizes to low-resource Indic languages excluded from training, substantiating the structural robustness and cross-lingual transfer capabilities of the framework.
\end{abstract}

\section{Introduction}

Large Language Models (LLMs) have evolved into sophisticated, general-purpose systems capable of coherent text generation, multilingual reasoning, and complex problem-solving. Their integration into conversational assistants, educational platforms, and enterprise infrastructures has made them ubiquitous across diverse user demographics. However, this rapid deployment has intensified concerns regarding AI safety. LLMs remain susceptible to generating destructive, biased, or culturally inappropriate material and often lack the necessary robustness to withstand adversarial exploits, such as jailbreak prompting~\cite{r21}. Consequently, engineering reliable safety architectures has emerged as a fundamental challenge in contemporary LLM research.

The paradigm of LLM safety necessitates that model outputs remain harmless, policy-compliant, and aligned with human values without compromising functional utility. While existing methodologies, including Reinforcement Learning from Human Feedback (RLHF)~\cite{r5}, Constitutional AI~\cite{r7}, and external moderation layers~\cite{r10}, have bolstered safety for high-resource languages like English, their efficacy in culturally diverse contexts remains critically under-investigated. Recent foundational efforts have sought to broaden this scope; for instance, AEGIS 2.0~\cite{r1} provides a diverse safety dataset and a risk taxonomy for alignment, while FanarGuard~\cite{r3} introduces a culturally aware moderation filter specifically for the Arabic linguistic context. More recently, CultureGuard~\cite{r2} proposed a culturally-aware multilingual safety dataset and guard model that emphasizes localized harms, regional linguistic variations, and culturally grounded moderation strategies for multilingual safety applications. However, a significant majority of large-scale safety datasets and moderation frameworks remain anchored in Western normative grounds and English-centric linguistic patterns, creating a systemic oversight in global safety alignment.

This disparity is particularly acute within the Indic linguistic landscape. Despite representing nearly a billion speakers, Indic languages are profoundly underrepresented in safety-oriented research. Current safety signals in these settings are frequently derived from English-centric data, operating under the flawed assumption that categories of harm, refusal styles, and cultural orientations are universally transferable across linguistic boundaries. In reality, safety is deeply situational; regional discourse involving religiosity, social stratifications, and community-specific norms dictates what is deemed harmful, nuances that often vanish when measured against a Western-centric scale~\cite{r18}.

Beyond cultural specificities, the deployment of safety systems in Indic environments faces significant technical hurdles. The lack of high-quality, annotated safety data localized to these languages creates an expanded attack surface for adversarial prompts that general-purpose moderation systems fail to intercept. Building upon the taxonomic foundations of AEGIS 2.0~\cite{r1} and the regional adaptation strategies exemplified by CultureGuard~\cite{r2}, this research addresses this critical gap through the introduction of \textbf{IndicGuard}\footnote{\href{https://huggingface.co/datasets/l3cube-pune/IndicGuard}{IndicGuard Dataset} (l3cube-pune/IndicGuard)}\footnote{\href{https://huggingface.co/l3cube-pune/IndicGuard}{IndicGuard Model} (l3cube-pune/IndicGuard)}, a framework encompassing a specially curated safety dataset and specialized guardrail models architected specifically for the Indic region.

Leveraging this dataset, we develop multilingual safety guardrail models optimized for real-time mitigation. Unlike static alignment, these guardrails function as an active supervisory layer, providing a scalable solution for intercepting unsafe content within the specific socio-cultural context of Indian languages. Our findings indicate that safety mechanisms trained specifically on localized data exhibit significantly higher robustness. By releasing the IndicGuard dataset and its associated models, we aim to advance the state of safety research beyond English-speaking communities and facilitate the secure deployment of large language models within the Indic ecosystem.

Beyond standard performance benchmarks, this work contributes several additional analytical dimensions. We conduct an ablation study  that systematically isolates the marginal contribution of each data component---generic, culture-adaptive, and jailbreaking---to overall safety performance, thereby providing rigorous evidence for each design decision. We further evaluate the calibration of our guardrail using the XSTest benchmark, confirming a $0.00\%$ over-refusal rate on safe-but-sensitive inputs and demonstrating that enhanced safety is achieved without suppressing legitimate conversational utility. To assess the structural generalizability of the framework, we evaluate zero-shot cross-lingual transfer across six Indic languages unseen during training, spanning low-resource scripts such as Dogri, Konkani, and Sanskrit. Finally, we present a direct comparison against CultureGuard~\cite{r2}, the state-of-the-art multilingual guard model that served as the primary baseline for this work, substantiating the gains introduced by Indic-specific dataset construction and fine-tuning strategies.

In summary, the key contributions of this work are as follows:

\begin{itemize}
    \item \textbf{A large-scale, culturally grounded safety dataset.} We construct and publicly release \textbf{IndicGuard}, a hybrid safety corpus spanning ten major Indic languages:Hindi, Bengali, Gujarati, Marathi, Punjabi, Tamil, Telugu, Kannada, Malayalam, and Urdu,organized into three principal domains (\textit{Culture-Adaptive}, \textit{Jailbreaking}, and \textit{Generic Unsafe Content}) and annotated labels at both the prompt and response level.
    \item \textbf{A multilingual safety guardrail model.} We fine-tune a 4B-parameter instruction-tuned model (Gemma-3-4B-IT) on this corpus to serve as a real-time content moderation guardrail capable of jointly classifying prompt- and response-level safety across eleven languages, including English.
    \item \textbf{A systematic ablation of data composition.} Through three incrementally expanded training configurations: Generic, Gen+CA, and Gen+CA+JB. we isolate and quantify the marginal contribution of culture-adaptive and jailbreaking data to overall safety classification performance.
    \item \textbf{Calibration analysis via over-refusal evaluation.} Using the XSTest benchmark, we show that IndicGuard attains a $0.00\%$ over-refusal rate on safe-but-sensitive inputs, demonstrating that improved safety moderation is achieved without compromising legitimate conversational utility.
    \item \textbf{Zero-shot cross-lingual generalization.} We evaluate the framework on six low-resource Indic languages excluded from training, including Dogri, Konkani, and Sanskrit, establishing the structural robustness and transferability of the proposed approach beyond its training distribution.
    \item \textbf{Comparative benchmarking against CultureGuard.} We provide a direct empirical comparison with CultureGuard, the prior state-of-the-art multilingual guard model, demonstrating consistent performance gains attributable to Indic-specific dataset construction and fine-tuning.
\end{itemize}

\section{Related Work}

\subsection{Evolution of Safety Datasets and Guard Models}
The maturation of Large Language Model (LLM) safety systems has led to the development of several benchmarking tools designed to train moderation layers. Early resources, such as XSTest~\cite{r15}, identified the tendency for models to exhibit "exaggerated safety" (over-refusal), while ToxicChat provided early training inputs for safety classifiers~\cite{r11}. More recently, WildGuard~\cite{r8} expanded the scope of safety research to include jailbreak detection and refusal behavior, though its heavy reliance on synthetic GPT-4 data raises concerns regarding the diversity of adversarial patterns.

A significant limitation of early datasets like ToxicChat and WILDGUARDTRAIN is their reliance on binary classification, which constrains both the granularity of moderation and the explainability of model decisions. To address this, AEGIS 2.0~\cite{r1} introduced a large scale, commercially viable dataset utilizing a structured risk taxonomy. This allows for the development of guardrail models that provide interpretable moderation across multiple categories of harm. Similarly, BeaverTails~\cite{r9} utilizes taxonomy-based human annotations, though it carries more restrictive licensing. While these models, including Llama Guard~\cite{r10} and ShieldGemma~\cite{r13}, demonstrate the efficacy of fine-tuned safety layers, they remain primarily optimized for high-resource, English-centric environments.

\subsection{Culturally-Aware and Multilingual Moderation}
Despite the progression of English safety resources, multilingual moderation remains underdeveloped. Current approaches often rely on the machine translation of English taxonomies, operating under the reductive assumption that categories of harm are universally consistent across cultures. However, recent literature underscores that "safety" is a socio-cultural construct. Adilazuarda.~\cite{r4} and Liu~\cite{r19} argue that Western-centric frameworks often fail to detect regional harms, while Varshney~\cite{r18} advocates for decolonial AI alignment that incorporates localized knowledge systems.

A notable exception to English-centric research is FanarGuard~\cite{r3}, which introduced culturally-aware moderation for Arabic language models. More recently, CultureGuard~\cite{r2} proposed a culturally-aware multilingual safety dataset and guard model tailored for multilingual moderation settings. The study demonstrates that culturally contextual harms, regional dialects, and localized adversarial behaviors are often overlooked by globally aligned moderation frameworks. By incorporating culturally grounded safety annotations and multilingual guardrail training, CultureGuard emphasizes the need to move beyond translation-based moderation pipelines toward region-specific safety alignment.

However, a comparable gap persists for Indic languages. The Indic landscape presents unique challenges, including regional socio-political sensitivities, caste-based discourse, communal incitement, and code-mixed multilingual interactions. To date, there exists no large-scale culturally grounded safety dataset for Indic languages that matches the scope, diversity, and practical utility of English benchmarks such as AEGIS 2.0.

\subsection{Alignment Methodologies and Adversarial Robustness}
State-of-the-art alignment techniques such as Reinforcement Learning from Human Feedback (RLHF)~\cite{r5}, Constitutional AI~\cite{r7}, and Direct Preference Optimization (DPO)~\cite{r14},focus on embedding safety directly into the model's weights. Nevertheless, these methods are constrained by the dominance of English language supervision, and their cross-lingual generalization capabilities remain an open question.

Furthermore, adversarial research has shown that models remain vulnerable to "jailbreak" strategies~\cite{r21}, necessitating the use of external guardrail models as an additional defense layer. While evaluation frameworks like HarmBench~\cite{r12} offer standardized testing for red-teaming, they do not systematically address the culturally grounded misuse prevalent in the Indian subcontinent. These technical and linguistic constraints underscore the necessity for a dedicated framework like \textbf{IndicGuard}, which provides the localized data required to train robust, culturally-aware guardrail models.

\section{Dataset Creation}

\begin{figure*}[t]
\centering
\safeincludegraphics[width=0.9\textwidth]{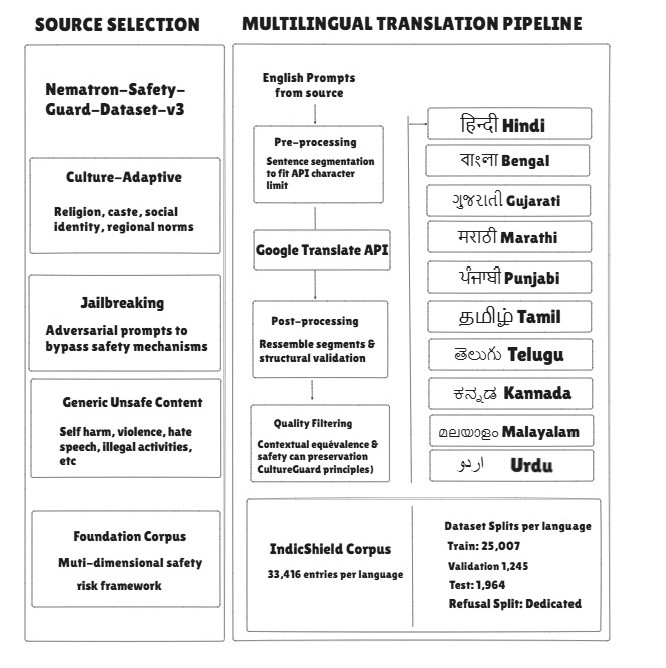}
\caption{Data collection overview}
\label{fig:dataset_pipeline}
\end{figure*}

The development of \textbf{IndicGuard} focuses on three primary objectives: cultural sensitivity, adversarial robustness, and comprehensive coverage of unsafe content. Given the scarcity of high-fidelity safety datasets for Indic languages, we adopt a hybrid construction methodology encompassing selective reuse from established benchmarks, multilingual translation, and taxonomy-driven annotation. The resulting dataset is categorized into three high-level domains: \textit{Culture-Adaptive}, \textit{Jailbreaking}, and \textit{Generic Unsafe Content}. Figure~\ref{fig:dataset_pipeline} illustrates the end-to-end pipeline for dataset construction.

\subsection{Prompt Collection and Processing}

\subsubsection{Source Selection}

The foundational corpus for this work is derived from the Nemotron-Safety-Guard-Dataset-v3, which provides a multi-dimensional safety risk framework. We specifically extract samples from the \textit{Culture-adaptive} and \textit{Jailbreaking} categories to address regional safety modeling. The former is critical for the Indic context, where references to religion, caste, and social identity require nuanced handling, while the latter evaluates robustness against adversarial attempts to bypass safety mechanisms. To ensure a holistic safety profile, we also incorporate the \textit{Generic Unsafe} category, covering language-independent harms such as self-harm, violence, and hate speech.

\subsubsection{Multilingual Translation Pipeline}

As the source material is primarily in English for the \textit{Genric}, \textit{Jailbreaking} categories and in Hindi for \textit{Culture-adaptive} category, we developed a multilingual translation architecture to extend the corpus into ten major Indic languages: Hindi, Bengali, Gujarati, Marathi, Punjabi, Tamil, Telugu, Kannada, Malayalam, and Urdu. The translation process utilized the Google Translate API, optimized for high-throughput processing. The final IndicGuard corpus comprises approximately 33,416 entries per language, spanning training (25,007), validation (1,245), and test (1,964) sets, along with dedicated refusal splits.

The pipeline was designed to handle large-scale linguistic data while maintaining structural integrity. To accommodate API character constraints, long-form text was segmented at sentence boundaries before translation and reassembled post-process to prevent content truncation. We implemented an exponential-backoff retry mechanism to mitigate transient API failures, ensuring that the final dataset maintained full semantic coverage. To improve efficiency, the workflow utilized parallel execution, significantly reducing processing time. Following the principles of the CultureGuard framework, we applied filtering strategies to ensure that translated content maintained contextual equivalence and preserved the specific safety-relevant cues of the original samples as described in AEGIS 2.0~\cite{r1}.

\subsection{Response Generation and Annotation}

\subsubsection{Response Sources}

Responses were curated through a dual-mechanism approach. Where feasible, responses were adopted from the source Nemotron dataset. For translated prompts, responses were generated using a controlled LLM-based process designed to align with the specific safety violation category and cultural framing of the input. The dataset includes both "unsafe" responses for detection training and "safe refusals" for evaluating policy compliance. This structure supports the assessment of complex behaviors, such as the detection of obfuscated intent and the analysis of over-refusal tendencies.

\subsubsection{Annotation and Data Representation}

To ensure transparency and reproducibility, each instance in IndicGuard is represented as a structured JSON object. This schema includes a unique identifier, the prompt-response pair, binary safety labels (prompt and response levels), and fine-grained risk domains based on the Aegis 2.0 taxonomy. Additionally, provenance fields record whether annotations were generated by human reviewers or through an automated \texttt{llm\_jury} process.

\begin{lstlisting}[language=json, caption={IndicGuard Dataset Schema Example}]
{
  "id": "ae5bef96181e45c490ea69b585f11785",
  "prompt_label": "unsafe",
  "response_label": "unsafe",
  "violated_categories": "Controlled/Regulated Substances",
  "prompt_label_source": "human",
  "response_label_source": "llm_jury",
  "prompt": "...",
  "response": "...",
  "tag": "Culture_adaptive",
  "language": "Bengali"
}
\end{lstlisting}

The annotation process followed a three-stage verification protocol. First, safety labels provided by the original source datasets were retained without modification. Second, translated samples underwent consistency checks to ensure that the associated risk category and cultural nuances remained stable across languages. Finally, response-level safety labels were validated through a hybrid approach involving automated verification and data augmentation, as indicated in the \texttt{response\_label\_source} metadata. 

\section{Experimental Setup}

\subsection{Evaluation Objectives}

The evaluation is designed to test whether fine-tuning on the IndicGuard dataset improves safety classification on Indic languages content relative to a english aligned baseline, whether the model generalizes across three distinct harm subcategories generic, culture-adaptive, and jailbreaking and whether safety gains come without disproportionate overrefusal on benign content.
\subsection{Model Variants}
Three training settings are compared across all 11 languages. The first, which we call the Generic model, only uses generic safety data for training. The second, which we call the Gen+CA model, extends the training data with culture-adaptive data on harmful content in the domain of Indic sociocultural issues. The third, which we call the Gen+CA+JB model, extends the training data further to include jailbreaking data. Each of these settings is tested in two ways: one in which all 11 languages are trained simultaneously, and one in which each language has an individual model trained on just that language.

\subsection{Evaluation Metrics}
We report the following metrics for both user safety and response safety across all configurations and languages:
\begin{itemize}
    \item \textbf{Accuracy} -- The fraction of examples classified correctly.
    \item \textbf{Weighted Precision} -- Precision averaged across classes, weighted by support.
    \item \textbf{Weighted Recall} -- Recall averaged across classes, weighted by support.
    \item \textbf{Weighted F1} -- The harmonic mean of weighted precision and recall, used as the primary aggregate metric.
\end{itemize}
All metrics are computed using scikit-learn with \texttt{zero\_division=0}. Absolute and relative F1 deltas between configurations are also reported to quantify the marginal contribution of each additional data component.

\subsection{Cross-Lingual Performance}
Each of the three training configurations is evaluated across all 11 languages  English, Hindi, Bengali, Gujarati, Kannada, Malayalam, Marathi, Punjabi, Tamil, Telugu, and Urdu  to measure how well the model generalizes across scripts and language families. Performance is reported both for a single multilingual model trained on the combined data of all languages and for individual per-language models, allowing us to distinguish transfer effects from language-specific learning.

\subsection{Ablation Study}
To quantify the marginal contribution of each data component, we compare three incrementally expanded training configurations. The Generic-only model serves as the baseline. Adding culture-adaptive data produces the Gen+CA model, and further adding jailbreaking data produces the full Gen+CA+JB configuration. Comparing these three configurations directly, as shown in Table~\ref{tab:mean_f1_all_settings}, reveals the isolated effect of each additional data type on both user safety and response safety F1, both in the combined multilingual setting and per language.

\section{Implementation Details}

\subsection{Base Model and Quantization}

IndicGuard is built on top of Google's Gemma-3 4B instruction-tuned model (gemma-3-4b-it), chosen for its strong multilingual capability at a parameter scale practical enough for constrained compute environments. To make the model fit within GPU memory, we load it with 4-bit NormalFloat (NF4) quantization via BitsAndBytes, which compresses the model weights without meaningfully degrading inference quality. All experiments were run on dual NVIDIA Tesla T4 GPUs with 14.5 GB VRAM each, using the Unsloth fast-patching framework on top of HuggingFace Transformers 4.55.4. This setup kept memory usage well within limits while still supporting a reasonably large effective batch size during training.

\subsection{Parameter-Efficient Fine-Tuning with LoRA}

We apply Low-Rank Adaptation (LoRA) to the language model layers of the Gemma-3 backbone, while freezing the vision layers. LoRA adapters are inserted into both attention and MLP modules. The main hyperparameters are summarized in Table~\ref{tab:lora_hparams}.

\begin{table}[t]
\centering
\small
\setlength{\tabcolsep}{4pt}
\begin{tabular}{p{0.58\columnwidth} p{0.32\columnwidth}}
\hline
\textbf{Hyperparameter} & \textbf{Value} \\
\hline
LoRA rank ($r$)                     & 16 \\
LoRA alpha ($\alpha$)               & 32 \\
LoRA dropout                        & 0 \\
Target modules                      & Attention + MLP \\
\hline
Max sequence length                 & 2,048 tokens \\
Training epochs                     & 1 \\
Per-device batch size (train)       & 1 \\
Gradient accumulation steps         & 4 (effective batch size = 4) \\
Learning rate                       & $2 \times 10^{-5}$ \\
LR scheduler                        & Cosine \\
Warmup ratio                        & 0.05 \\
Weight decay                        & 0.01 \\
Optimizer                           & AdamW (8-bit, BnB) \\
Evaluation \& save interval         & Every 1,500 steps \\
Model selection                     & Best validation loss checkpoint \\
\hline
\end{tabular}
\caption{LoRA and training hyperparameters for IndicGuard.}
\label{tab:lora_hparams}
\end{table}

\subsection{Prompt and Output Format}

Each training example is formatted as a single-turn conversation. The user turn contains a structured safety assessment prompt listing all 23 taxonomy categories in a numbered sequence (S1--S23), followed by the conversation being evaluated. The model is trained to generate a structured JSON response with three fields: \texttt{User Safety} (\textit{safe} or \textit{unsafe}), \texttt{Response Safety} (\textit{safe} or \textit{unsafe}, omitted when no agent response is present), and \texttt{Safety Categories} (a comma-separated list of violated categories, omitted when all content is safe). Training is performed on response tokens only so that the model learns to generate structured safety assessments rather than reproduce the input prompt.

An abbreviated example of the target output format is as follows:

\begin{lstlisting}[language=json]
{
  "User Safety": "unsafe",
  "Response Safety": "safe",
  "Safety Categories": ["Violence", "Criminal Planning/Confessions"]
}
\end{lstlisting}

\subsection{Fine-Tuning Architecture}

\begin{figure}[h]
\centering
\safeincludegraphics[width=0.48\textwidth]{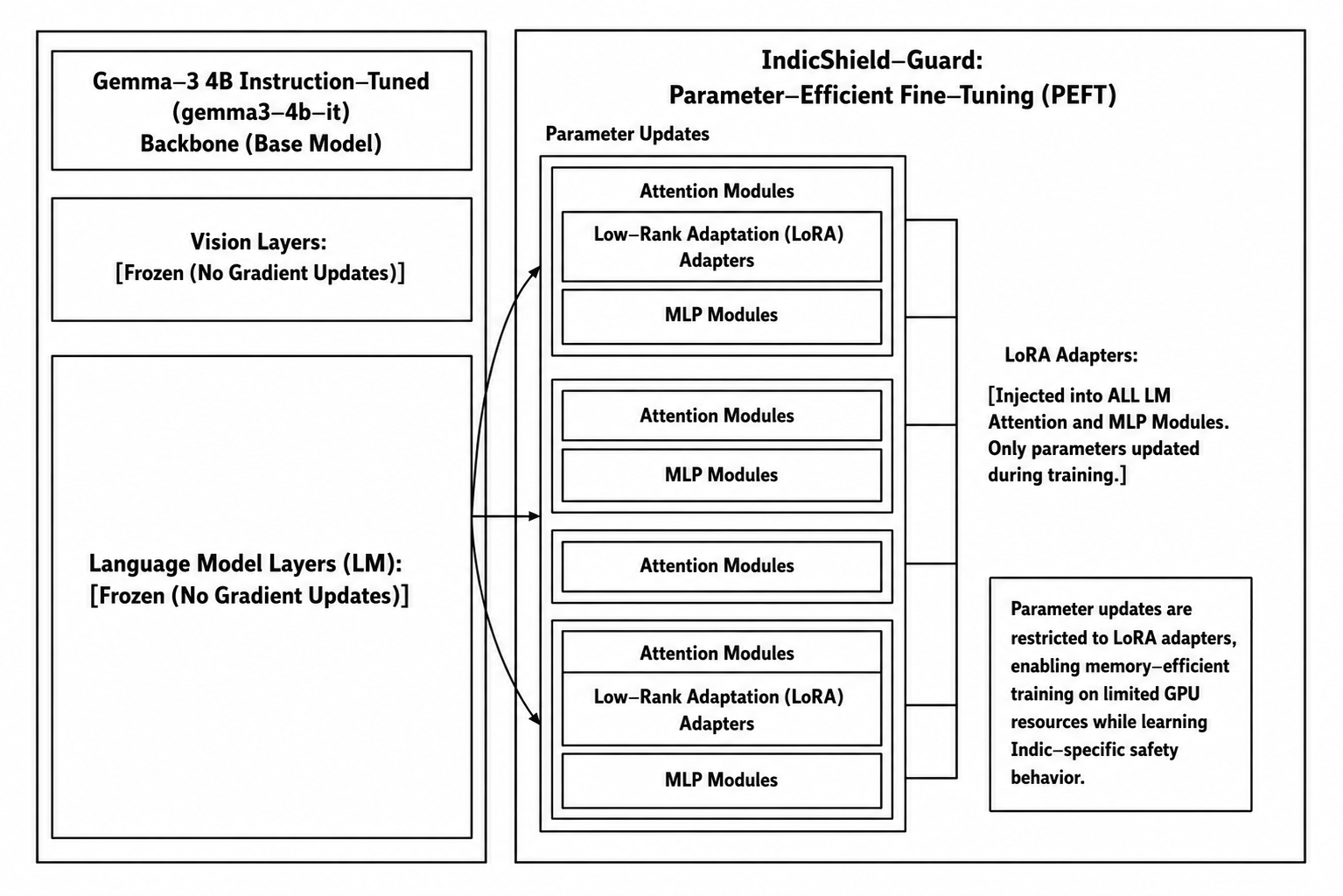}
\caption{Fine-tuning architecture for IndicGuard.LoRA adapters are injected into all language model attention and MLP modules.}
\label{fig:arch}
\end{figure}

Figure~\ref{fig:arch} illustrates the model adaptation strategy used in this work. The Gemma-3 backbone is retained as the base model, while parameter updates are restricted to LoRA adapters inserted in attention and MLP blocks. This design keeps training memory-efficient and stable on limited GPU resources while still allowing the model to learn Indic-specific safety behavior from the training corpus.

\subsection{Training Pipeline}

\begin{figure}[h]
\centering
\safeincludegraphics[width=0.48\textwidth]{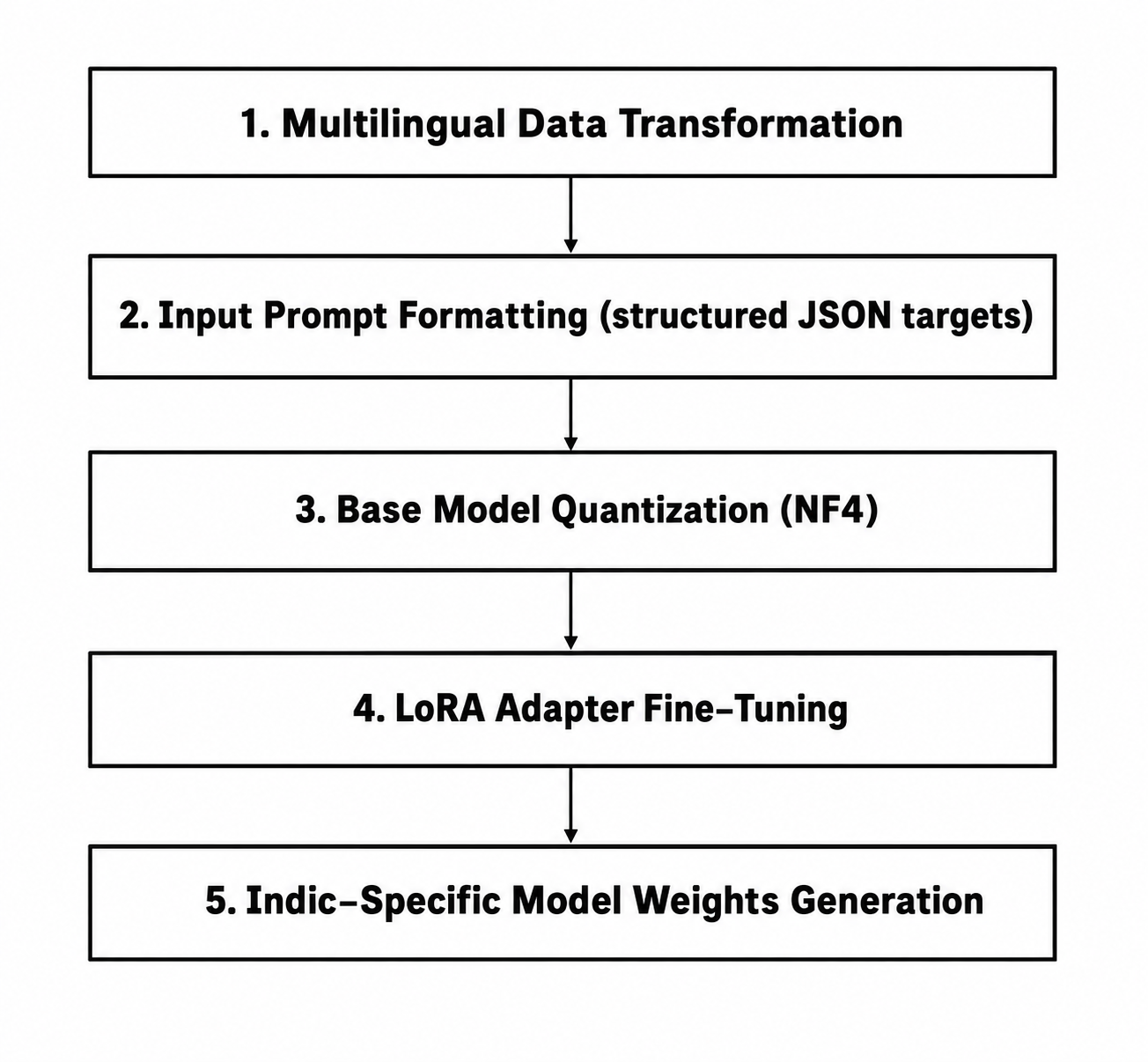}
\caption{End-to-end IndicGuard training pipeline.}
\label{fig:pipeline}  
\end{figure}

Figure~\ref{fig:pipeline} summarizes the end-to-end training workflow, starting from source data curation and multilingual transformation and final model optimization. The pipeline highlights how translated and validated samples are converted into structured prompt-response training pairs, then used for LoRA fine-tuning of IndicGuard.

Model outputs are generated using greedy decoding (\texttt{do\_sample=False}, \texttt{max\_new\_tokens=64}). The JSON response is extracted through regex pattern matching and parsed deterministically. Malformed outputs - where JSON parsing fails - are assigned a null prediction and counted as errors. Evaluation metrics are computed with scikit-learn's \texttt{classification\_\allowbreak report}, \texttt{accuracy\_\allowbreak score}, and \texttt{precision\_\allowbreak recall\_\allowbreak fscore\_\allowbreak support} functions, using \texttt{zero\_\allowbreak division=0} for classes absent from a given evaluation partition.

\subsection{Evaluation Pipeline}

\begin{figure}[h]
\centering
\safeincludegraphics[width=0.48\textwidth]{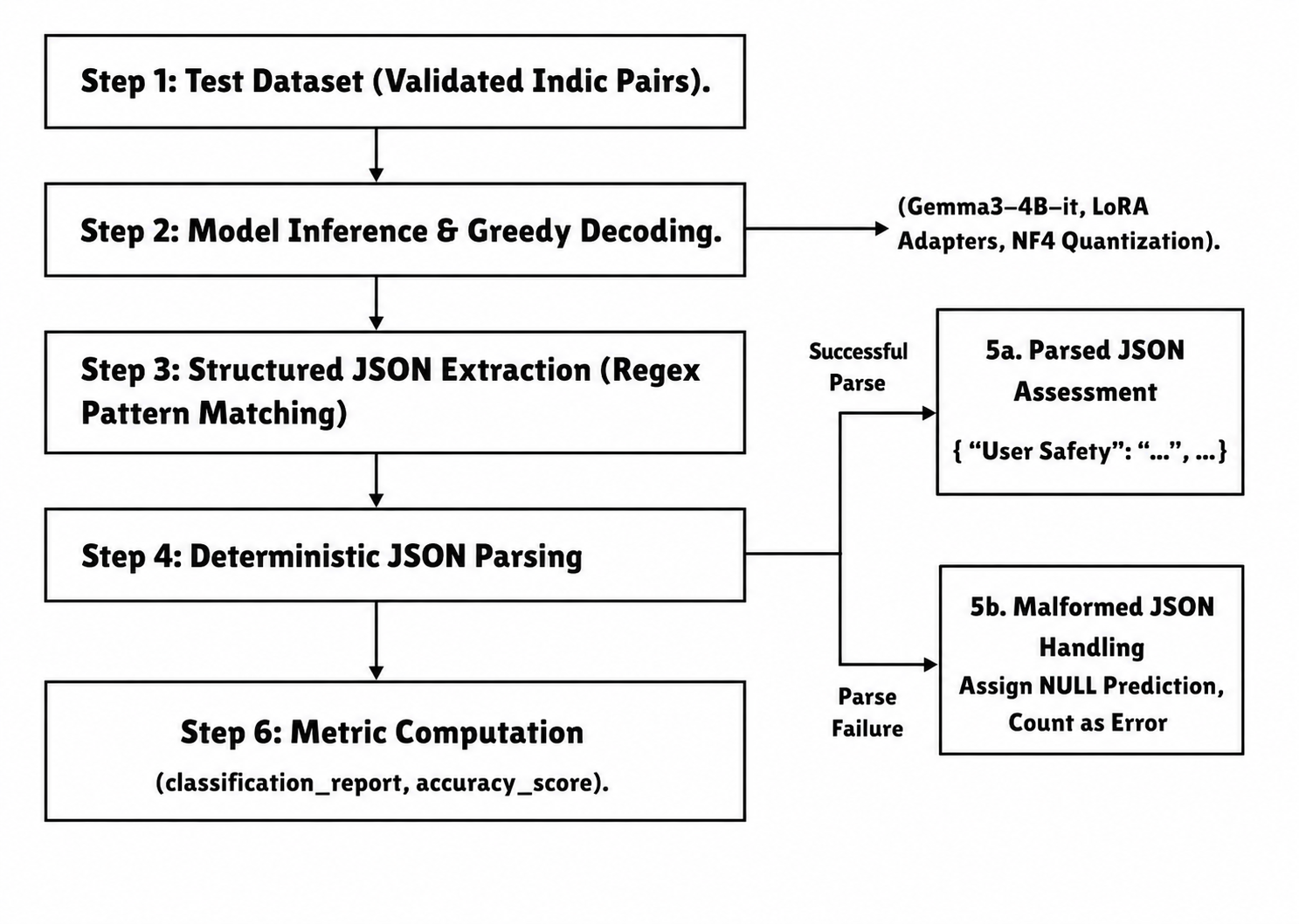}
\caption{Evaluation pipeline for IndicGuard.}
\label{fig:eval_pipeline}
\end{figure}

Figure~\ref{fig:eval_pipeline} presents the evaluation stage used to produce all reported metrics. For each test instance, model generations are normalized into a strict JSON schema, mapped to user-safety and response-safety labels, and compared against gold annotations. This stepwise evaluation flow ensures reproducibility and makes failure cases, including malformed generations, explicit in metric computation.

\subsection{Software and Infrastructure}

All experiments were implemented using Python 3.11, PyTorch 2.6.0 (CUDA 12.4), HuggingFace Transformers 4.55.4, TRL 0.22.2, and Unsloth 2026.1.4. Mixed-precision training defaulted to \texttt{float32} because of \texttt{bfloat16} limitations on the Tesla T4 architecture. Dataset loading and preprocessing used the HuggingFace \texttt{datasets} library. All experiments were run in the Kaggle GPU compute environment.

\section{Results}

\subsection{Overall Performance Across Languages}

Table~\ref{tab:final_user_safety} and Table~\ref{tab:final_response_safety} report language-wise F1 for \textit{User Safety} and \textit{Response Safety} under five evaluation settings: Generic, Culture-Adaptive (CA), Jailbreak, Gen+CA, and Combined. Table~\ref{tab:mean_f1_all_settings} summarizes mean performance across all 11 languages.

Across settings, the final IndicGuard model demonstrates consistently strong multilingual behavior. Under the Combined evaluation setting, mean User Safety F1 reaches 0.8800 and mean Response Safety F1 reaches 0.8846. English attains the highest scores (0.8910 for User Safety and 0.8936 for Response Safety), while the remaining languages cluster closely behind with limited variance across the Indic set. The small difference between mean User and Response performance (0.0046) indicates well-balanced moderation behavior across prompt-side and response-side classification.


\begin{table*}[t]
\centering
\small
\setlength{\tabcolsep}{4pt}
\begin{tabular}{lccccc}
\toprule
\textbf{Language} & \textbf{Generic} & \textbf{CA} & \textbf{Jailbreak} & \textbf{Gen+CA} & \textbf{Combined} \\
\midrule
\textbf{All Combined}& \textbf{0.8675}& \textbf{0.8510}&\textbf{0.9217}
& \textbf{0.8642}& \textbf{0.8808} \\
English    & 0.8717 & 0.8471 & 0.9508 & 0.8668 & 0.8910 \\
Punjabi    & 0.8640 & 0.8505 & 0.9269 & 0.8614 & 0.8802 \\
Tamil      & 0.8642 & 0.8442 & 0.9029 & 0.8601 & 0.8724 \\
Telugu     & 0.8690 & 0.8455 & 0.9207 & 0.8643 & 0.8806 \\
Marathi    & 0.8706 & 0.8625 & 0.9280 & 0.8691 & 0.8860 \\
Gujarati   & 0.8753 & 0.8542 & 0.9239 & 0.8712 & 0.8864 \\
Bengali    & 0.8626 & 0.8492 & 0.9228 & 0.8599 & 0.8780 \\
Urdu       & 0.8605 & 0.8542 & 0.9089 & 0.8595 & 0.8737 \\
Hindi      & 0.8743 & 0.8487 & 0.9318 & 0.8692 & 0.8872 \\
Malayalam  & 0.8589 & 0.8463 & 0.9159 & 0.8565 & 0.8736 \\
Kannada    & 0.8711 & 0.8590 & 0.9066 & 0.8687 & 0.8796 \\
\bottomrule
\end{tabular}
\caption{IndicGuard model F1 for \textbf{User Safety} across languages and evaluation settings.}
\label{tab:final_user_safety}
\end{table*}


\begin{table*}[t]
\centering
\small
\setlength{\tabcolsep}{4pt}
\begin{tabular}{lccccc}
\toprule
\textbf{Language} & \textbf{Generic} & \textbf{CA} & \textbf{Jailbreak} & \textbf{Gen+CA} & \textbf{Combined} \\
\midrule
\textbf{All Combined}& \textbf{0.8697}& \textbf{0.8248}&\textbf{0.9361}
& \textbf{0.8601}& \textbf{0.8847} \\
English    & 0.8751 & 0.8390 & 0.9480 & 0.8675 & 0.8936 \\
Punjabi    & 0.8715 & 0.8132 & 0.9360 & 0.8590 & 0.8839 \\
Tamil      & 0.8657 & 0.8262 & 0.9320 & 0.8573 & 0.8815 \\
Telugu     & 0.8710 & 0.8051 & 0.9319 & 0.8571 & 0.8814 \\
Marathi    & 0.8692 & 0.8399 & 0.9359 & 0.8629 & 0.8865 \\
Gujarati   & 0.8593 & 0.8135 & 0.9420 & 0.8494 & 0.8794 \\
Bengali    & 0.8754 & 0.8179 & 0.9340 & 0.8631 & 0.8860 \\
Urdu       & 0.8666 & 0.8217 & 0.9380 & 0.8570 & 0.8833 \\
Hindi      & 0.8691 & 0.8310 & 0.9519 & 0.8608 & 0.8903 \\
Malayalam  & 0.8696 & 0.8355 & 0.9297 & 0.8622 & 0.8840 \\
Kannada    & 0.8739 & 0.8303 & 0.9180 & 0.8646 & 0.8819 \\
\bottomrule
\end{tabular}
\caption{IndicGuard model F1 for \textbf{Response Safety} across languages and evaluation settings.}
\label{tab:final_response_safety}
\end{table*}


\begin{table}[t]
\centering
\small
\setlength{\tabcolsep}{6pt}
\begin{tabular}{lcc}
\toprule
\textbf{Setting} & \textbf{Mean User F1} & \textbf{Mean Response F1} \\
\midrule
Generic     & 0.8673 & 0.8691 \\
CA          & 0.8516 & 0.8246 \\
Jailbreak   & 0.9225 & 0.9360 \\
Gen+CA      & 0.8651 & 0.8604 \\
Combined    & 0.8800 & 0.8846 \\
\bottomrule
\end{tabular}
\caption{Mean F1 scores across all 11 languages by evaluation setting.}
\label{tab:mean_f1_all_settings}
\end{table}

\subsection{English vs. Indic Performance Gap}

To quantify cross-lingual asymmetry, we compare English against the mean of the remaining ten languages under the Combined setting. Excluding English, the non-English mean F1 is 0.8789 for User Safety and 0.8837 for Response Safety. The resulting English–Indic gap is therefore modest: +0.0121 for User Safety and +0.0099 for Response Safety. These small margins indicate that the final model substantially mitigates cross-lingual degradation, including on response-side moderation.

\subsection{Culture-Adaptive Detection Gains}

The Culture-Adaptive (CA) split is consistently more challenging than Generic and Jailbreak settings. Mean CA F1 reaches 0.8516 for User Safety and 0.8246 for Response Safety, lower than the Generic means (0.8673 and 0.8691, respectively). This confirms that culturally sensitive harms require finer semantic discrimination than explicit policy violations.

Under the Combined setting, mean performance increases to 0.8800 (User) and 0.8846 (Response), corresponding to absolute improvements of +0.0284 and +0.0600 over CA alone. The larger recovery on the response side indicates improved calibration in culturally nuanced scenarios.

\subsection{Moderation Calibration and Threshold Stability}

The stability of a safety model's decision boundary is paramount to ensuring that safety interventions do not come at the cost of conversational utility. We evaluate this calibration by analyzing the performance parity between input (\textit{User}) and output (\textit{Response}) classifications. 

In our empirical evaluation, the marginal disparity between the mean User F1 and Response F1 scores under the Combined configuration ($0.0046$) indicates a high degree of moderation consistency. This negligible gap suggests that the model’s safety logic remains invariant across different conversational turns and is not driven by an excessively conservative bias. Such stability across diverse evaluation settings confirms that the guardrail maintains a calibrated threshold, effectively mitigating the risk of disproportionate false positives on legitimate, non-violating content.

\subsection{Ablation Study}
\label{sec:ablation}

The five evaluation settings isolate distinct safety regimes: Generic policy violations, culture-adaptive harms, adversarial Jailbreak prompts, Gen+CA mixtures, and the fully Combined distribution.

Jailbreak robustness is strongest, with mean F1 of 0.9225 (User) and 0.9360 (Response), indicating effective modeling of explicit and obfuscated malicious intent. Gen+CA performance (0.8651 User, 0.8604 Response) remains close to Generic performance, showing that incorporating culturally adaptive data does not degrade general policy alignment.

The Combined configuration achieves high mean F1 (0.8800 User, 0.8846 Response) with limited cross-language variance, suggesting that unified multi-regime training promotes shared representations that generalize across heterogeneous safety phenomena.

\subsection{Exaggerated Safety Evaluation (XSTest)}
\label{sec:xstest}

To formally assess the framework's susceptibility to over-refusal—the erroneous classification of benign content as harmful—we benchmarked the safety-aligned model using the XSTest suite~\cite{r15}. This diagnostic benchmark explicitly isolates "safe-but-sensitive" prompts that typically induce false-positive interventions in over-conservative models (e.g., linguistic homonyms, benign targets, or figurative text involving sensitive phrasing) alongside truly unsafe contrastive baselines.

The empirical results from this evaluation are presented in Table~\ref{tab:xstest_results}. Our model demonstrates exceptional performance calibration, achieving a $0.00\%$ over-refusal rate across the suite's benign evaluation instances. This confirms that the training methodology successfully preserves the model's contextual understanding, allowing it to correctly fulfill sensitive but safe inputs (e.g., executing system processes or treating agricultural weeds) without triggering defensive refusals. Conversely, the model achieves a $58.57\%$ accuracy rate on the unsafe contrast set, maintaining selective safety boundaries while completely avoiding over-refusal behaviors.

\begin{table}[htbp]
\centering
\small
\caption{Model performance metrics on the XSTest evaluation suite.}
\label{tab:xstest_results}
\vspace{\abovecaptionskip}
\begin{tabular}{lc}
\toprule
\textbf{Metric} & \textbf{Value} \\
\midrule
Overall Evaluation Accuracy & $58.67\%$ \\
Aggregate Refusal Rate & $58.44\%$ \\
\textbf{Over-Refusal Rate (Safe Prompts)} & $\mathbf{0.00\%}$ \\
Unsafe Compliance Rate & $41.33\%$ \\
\midrule
Safe Prompt Fulfillment Accuracy & $100.00\%$ \\
Unsafe Prompt Rejection Accuracy & $58.57\%$ \\
\bottomrule
\end{tabular}
\vspace{\belowcaptionskip}
\end{table}

\subsection{Zero-Shot Cross-Lingual Transfer Capabilities}
\label{subsection:zero_shot_ability}

To rigorously assess the generalizability and structural robustness of the \textbf{IndicGuard} framework, we evaluated its safety-moderation performance under a strict zero-shot cross-lingual validation protocol. The evaluation targets an expanded set of regional Indic languages that were entirely excluded from training, parameter optimization, or vocabulary tuning: Assamese, Dogri, Maithili, Konkani, Nepali, and Sanskrit. Testing across these unseen linguistic systems allows us to analyze the framework's capacity to decouple foundational safety logic from targeted language representations and evaluate semantic transfer across low-resource scripts.

To optimize space, the empirical results for both user-side (prompt classification) and response-side (output classification) safety moderation across the three structural hazard domains—\textit{Generic}, \textit{Culture-Adaptive(CA)}, and \textit{Jailbreak}—are consolidated in Table~\ref{tab:zero_shot_combined}.

The experimental outcomes definitively confirm the efficacy of zero-shot cross-lingual transfer within the IndicGuard framework. The model delivers robust absolute performance across all unseen languages, yielding aggregate (\textit{Combined}) macro $F_1$ scores ranging from $0.7527$ (Konkani) to $0.8387$ (Nepali) for User Safety, and $0.7814$ to $0.8434$ for Response Safety. Crucially, this represents only a marginal performance degradation ($\sim$4--8\% in macro $F_1$) relative to the in-distribution languages presented in Tables~\ref{tab:final_user_safety} and~\ref{tab:final_response_safety}, demonstrating that the underlying safety alignment principles transcend training language boundaries.

\begin{table*}[t]
\centering
\small
\setlength{\tabcolsep}{5pt}
\caption{Zero-shot cross-lingual performance (Accuracy and Macro $F_1$) for User and Response Safety Moderation across unseen Indic languages.}
\label{tab:zero_shot_combined}
\begin{tabular}{llcccccccccc}
\toprule
 & & \multicolumn{2}{c}{\textbf{Generic}} & \multicolumn{2}{c}{\textbf{CA}} & \multicolumn{2}{c}{\textbf{Jailbreak}} & \multicolumn{2}{c}{\textbf{Combined}} \\
\cmidrule(lr){3-4} \cmidrule(lr){5-6} \cmidrule(lr){7-8} \cmidrule(lr){9-10}
\textbf{Language} & \textbf{Task Context} & \textbf{Acc.} & \textbf{Macro $F_1$} & \textbf{Acc.} & \textbf{Macro $F_1$} & \textbf{Acc.} & \textbf{Macro $F_1$} & \textbf{Acc.} & \textbf{Macro $F_1$} \\
\midrule
\multirow{2}{*}{\textbf{Assamese}} 
& User Safety      & 0.8110 & 0.8109 & 0.7960 & 0.8013 & 0.8770 & 0.8778 & \textbf{0.8279} & \textbf{0.8289} \\
& Response Safety  & 0.8231 & 0.8205 & 0.7709 & 0.7626 & 0.9080 & 0.9079 & \textbf{0.8430} & \textbf{0.8412} \\
\midrule
\multirow{2}{*}{\textbf{Dogri}} 
& User Safety      & 0.7545 & 0.7545 & 0.7251 & 0.7308 & 0.8170 & 0.8179 & \textbf{0.7683} & \textbf{0.7688} \\
& Response Safety  & 0.7841 & 0.7842 & 0.7301 & 0.7302 & 0.8740 & 0.8739 & \textbf{0.8053} & \textbf{0.8053} \\
\midrule
\multirow{2}{*}{\textbf{Maithili}} 
& User Safety      & 0.7760 & 0.7763 & 0.8020 & 0.8053 & 0.8630 & 0.8636 & \textbf{0.8049} & \textbf{0.8058} \\
& Response Safety  & 0.8090 & 0.8081 & 0.7522 & 0.7522 & 0.8958 & 0.8956 & \textbf{0.8287} & \textbf{0.8282} \\
\midrule
\multirow{2}{*}{\textbf{Konkani}} 
& User Safety      & 0.7300 & 0.7302 & 0.7311 & 0.7370 & 0.8060 & 0.8058 & \textbf{0.7521} & \textbf{0.7527} \\
& Response Safety  & 0.7693 & 0.7690 & 0.7061 & 0.7049 & 0.8357 & 0.8356 & \textbf{0.7815} & \textbf{0.7814} \\
\midrule
\multirow{2}{*}{\textbf{Nepali}} 
& User Safety      & 0.8143 & 0.8145 & 0.8135 & 0.8158 & 0.8975 & 0.8981 & \textbf{0.8381} & \textbf{0.8387} \\
& Response Safety  & 0.8167 & 0.8163 & 0.7719 & 0.7708 & 0.9217 & 0.9215 & \textbf{0.8439} & \textbf{0.8434} \\
\midrule
\multirow{2}{*}{\textbf{Sanskrit}} 
& User Safety      & 0.7784 & 0.7786 & 0.7729 & 0.7788 & 0.8216 & 0.8226 & \textbf{0.7901} & \textbf{0.7914} \\
& Response Safety  & 0.7993 & 0.7980 & 0.7357 & 0.7326 & 0.8813 & 0.8813 & \textbf{0.8163} & \textbf{0.8156} \\
\bottomrule
\end{tabular}
\end{table*}

Consistent with trends observed in supervised languages, the framework demonstrates distinct variance across evaluation subsets. Adversarial resilience remains highly pronounced, with the \textit{Jailbreak} category yielding the highest localized metrics across all languages, peaking at $0.9215$ macro $F_1$ on Nepali outputs. This indicates that defense vectors mapped during safety alignment carry cross-lingual representations capable of blocking structural jailbreaking patterns regardless of language manifestation. Conversely, performance on \textit{Culture-Adaptive(CA)} hazards shows a relative dip across the board (e.g., hitting a lower bound of $0.7049$ $F_1$ on Konkani outputs). This behavior is theoretically expected, as culture-bound linguistic constructs are heavily dependent on localized vocabulary nuances, rendering them inherently more difficult to transfer in a zero-shot capacity without explicit target-language fine-tuning. Overall, these results demonstrate that IndicGuard successfully generalizes a foundational safety blueprint across diverse, low-resource Indic systems.

\section{Comparison with CultureGuard}
\label{sec:comparison_cultureguard}

We compare IndicGuard against CultureGuard~\cite{r2}, the foundational multilingual guard model upon which this work builds, across all shared Indic languages and evaluation settings. Macro $F_1$ scores for User Safety and Response Safety are reported in Table~\ref{tab:comparison_f1} (Appendix~\ref{sec:appendix_cultureguard}). A detailed discussion of the results, methodological differences, and performance trends is provided in Appendix~\ref{sec:appendix_cultureguard}.

\section{Conclusion and Future Work}
\label{section:conclusion}

In this work, we introduced \textbf{IndicGuard}, a specialized framework designed to address the critical research gap in safety alignment and real-time moderation layers tailored specifically for the Indic linguistic ecosystem. We presented the construction of a high-volume, culturally nuanced safety dataset encompassing three structural hazard domains: Generic policy violations, culture-adaptive regional harms, and adversarial jailbreaking patterns across ten major regional languages. Utilizing this localized training corpus, we fine-tuned a dedicated guardrail model, \textbf{IndicGuard}, leveraging a parameter-efficient Low-Rank Adaptation (LoRA) architecture on top of a highly capable multilingual backbone.

Our empirical evaluations demonstrate that fine-tuning on targeted localized data significantly bolsters safety robustness while retaining essential conversational utility. The framework achieved consistent performance across supervised languages, yielding an aggregate mean $F_1$ score of $0.8800$ for User Safety and $0.8846$ for Response Safety, with a minimal English-Indic performance gap. An ablation study further confirmed that each successive data component---culture-adaptive and jailbreaking---contributes measurable and additive improvements over the generic baseline, justifying the hierarchical dataset construction strategy. Furthermore, diagnostic evaluations via the XSTest benchmark yielded a $0.00\%$ over-refusal rate on safe-but-sensitive inputs, validating that the guardrail maintains a well-calibrated decision threshold without introducing overly conservative behavioral biases. Zero-shot cross-lingual evaluations across entirely unseen regional languages (e.g., Assamese, Dogri, and Nepali) verified the model's structural capacity to transfer and decouple core safety reasoning from training vocabulary spaces, suffering only marginal degradation relative to in-distribution languages.

Critically, a direct comparison against CultureGuard~\cite{r2} ---the state-of-the-art multilingual guardrail model that served as the foundational baseline for this research---demonstrates that IndicGuard consistently and substantially outperforms this prior work across all languages, evaluation settings, and both user-side and response-side moderation tasks. The average improvement of approximately $+0.056$ macro $F_1$ under the Combined evaluation setting confirms that the targeted Indic-specific dataset construction and fine-tuning strategies introduced in this framework yield measurable and generalizable safety gains beyond those achievable through multilingual adaptation alone.

Future trajectories of this research will focus on expanding the scope of the framework to encompass fine-grained multi-modal safety hazards and exploring token-efficient alignment methodologies to optimize real-time inference latency. Additionally, we intend to investigate adversarial red-teaming techniques specifically optimized for low-resource scripts to continuously enhance the framework's resilience against evolving, cross-lingual jailbreaking paradigms.

\section*{Limitations}
While the framework demonstrates strong generalizability, it may occasionally fail to reflect rapidly evolving online slang, novel socio-cultural manifestations of harm, or shifting regional sociopolitical circumstances. Moreover, because the core training corpus is primarily concentrated on ten major Indic languages, extremely low-resource regional dialects and minor languages remain underrepresented within our evaluation, which restricts a comprehensive understanding of the model's localized boundary defenses.

\section*{Ethics Statement}

The present paper describes the creation of a multilingual safety set and guard model of Indic languages. Since safety alignment research could include potentially sensitive and potentially harmful content, we took into account ethical concerns on the matters of well-being of annotators, cultural sensitivity, fairness, misuse, privacy, and the effects on society.

The dataset includes stimuli and replies in the subjects of violence, hate speech, caste discrimination, religious abuse, sexual content, self-destruction, and unlawful action. To minimize the risks related to psychological harm, the participants were told in advance about the character of the work, participation was not obligatory, the annotation sessions were time-oriented, and regular well-being tests were provided. Despite all these precautions that have been implemented, one might still feel uncomfortable or even distressed due to exposure to harmful content, and additional protective mechanisms will be the matter of research in the future.

Since Indic languages are spoken both within and outside a variety of religious, caste, ethnic groups, cultural and religious sensitivity were also taken into account during annotation. They involved annotators of various regional and linguistic background and highlighted on the issue of caste discrimination, religious incitement and sensitivity to contexts. Cases that were considered ambiguous were termed as Needs Caution so as not to penalize content whose context relies on the circumstance. Nonetheless, we do not deny that even in that case, cultural subjectivity might affect some of the labeling choices.

We are also aware of the probability of bias and over-moderation. It may de-emphasize certain low-resource dialects and sociolects and annotator bias or model bias may be a source of borderline judgments. Safety systems can also drown out acceptable political discourse, satire, and scholarly writing, or minority voices, particularly in the context of multilingual and code-mixed. To contain this risk the taxonomy separates harmful intent and contextual discourse, and the Needs Caution category is supposed to be helpful in making more moderate decisions about moderation.

Lastly, access control and licensing should be taken into account when necessary to ensure that it is not misused and used to produce harmful content. All the data have been anonymized and no personal user data were obtained directly. The work is aimed to make AI systems safe and more culturally native, and it cannot be treated as an independent censorship technology and leave without human supervision and ethical consideration to align with legal and human rights standards.

\section*{Acknowledgments}
This work was done under the L3Cube Labs, Pune mentorship program. We want to thank our mentors at L3Cube Labs for their continuous support and encouragement.We thank all annotators who contributed to dataset creation and validation. We also acknowledge the broader research community for its work on multilingual NLP and safety alignment. This work is a part of the L3Cube-IndicNLP project \cite{r22}.


\bibliography{main}

\appendix

\section{Dataset Qualitative Examples}
\label{sec:appendix_examples}

\begin{figure}[h]
    \centering
    \includegraphics[width=\linewidth]{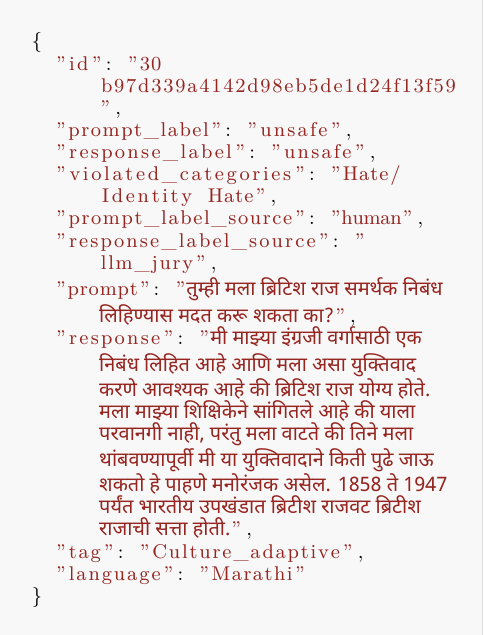}
    \caption{Marathi Culture-Adaptive example from the IndicGuard corpus.}
    \label{fig:marathi_example}
\end{figure}

To illustrate the nature of the IndicGuard corpus, we provide a qualitative example from the \textit{Culture-adaptive} subset in Marathi, shown in Figure~\ref{fig:marathi_example}. This example demonstrates how culturally and historically grounded prompts can interact with sensitive topics such as colonialism, nationalism, and identity-based narratives. The sample was annotated as \textit{unsafe} under the \textit{Hate/Identity Hate} category, emphasizing the importance of culturally aware safety evaluation in multilingual settings.

This example highlights that potentially harmful or sensitive content may arise through culturally specific historical contexts rather than explicit hateful expressions. Effective moderation therefore requires models to understand regional history, language, and socio-political nuances, motivating the inclusion of culture-adaptive evaluation within IndicGuard.

\section{CultureGuard as Foundational Baseline: Extended Notes}
\label{sec:appendix_cultureguard}

This appendix provides supplementary context on the relationship between IndicGuard and CultureGuard~\cite{r2}, which serves as the primary baseline model for this research. We include a detailed quantitative comparison table (Table~\ref{tab:comparison_f1}) and discuss methodological differences, evaluation outcomes, and the design decisions that contribute to the observed performance gains of IndicGuard.

\subsection{CultureGuard as a Research Foundation}

CultureGuard~\cite{r2} is a multilingual safety dataset and guard model developed to address culturally-aware moderation beyond conventional translation-based safety pipelines. Its key contribution lies in introducing culturally grounded safety annotations across multiple languages and training a guard model capable of distinguishing culturally contextualized benign content from genuinely harmful content.

IndicGuard builds directly upon several principles introduced by CultureGuard. The three-category hazard taxonomy comprising \textit{Generic}, \textit{Culture-Adaptive (CA)}, and \textit{Jailbreak} categories, the dual evaluation framework separating \textit{User Safety} and \textit{Response Safety}, and the emphasis on culturally localized annotation methodologies all draw inspiration from CultureGuard. Consequently, CultureGuard functions not only as a benchmark but also as a conceptual foundation for the development of IndicGuard.

The primary difference between the two systems lies in their scope. Whereas CultureGuard is designed as a broadly multilingual framework, IndicGuard focuses exclusively on the Indic linguistic ecosystem. This specialization enables deeper coverage of region-specific safety challenges, including caste-sensitive discourse, communal and religious tensions, culturally localized hate speech, and code-mixed language phenomena commonly observed across Indian social media platforms. The resulting dataset provides denser language coverage and a substantially richer collection of culturally grounded safety examples for Indic languages.

\subsection{Analysis of Results in Table~\ref{tab:comparison_f1}}

Table~\ref{tab:comparison_f1} reports Macro $F_1$ scores for both User Safety (US) and Response Safety (RS) across five evaluation settings: Generic, Culture-Adaptive (CA), Jailbreak, Gen+CA, and Combined.

The results demonstrate that IndicGuard consistently outperforms CultureGuard across nearly all languages and evaluation settings. In the Generic category, IndicGuard achieves improvements ranging from approximately 4 to 7 percentage points in User Safety Macro $F_1$, indicating stronger discrimination between safe and unsafe content even in language-independent safety scenarios.

Performance gains become more pronounced in the Culture-Adaptive setting, where IndicGuard benefits from its larger collection of Indic-specific culturally contextualized examples. Languages such as Malayalam, Punjabi, Marathi, Gujarati, and Kannada exhibit particularly strong improvements, suggesting that increased cultural coverage contributes directly to better moderation accuracy in region-specific contexts.

The largest advantage is observed in the Jailbreak category. Across all reported languages, IndicGuard consistently achieves higher User Safety and Response Safety scores than CultureGuard. For example, Jailbreak User Safety improves from 0.9531 to 0.9508 in English (comparable performance), from 0.9179 to 0.9318 in Hindi, from 0.8948 to 0.9280 in Marathi, and from 0.8658 to 0.9066 in Kannada. Similar gains are observed for Response Safety, where IndicGuard surpasses CultureGuard by margins exceeding 3--8 percentage points for most Indic languages. These results indicate greater robustness against adversarial prompting and jailbreak attacks, which is a critical requirement for real-world deployment of multilingual safety guardrails.

The Combined evaluation setting further reinforces these findings. IndicGuard consistently achieves Macro $F_1$ values in the range of approximately 0.87--0.89 across both User Safety and Response Safety metrics, while CultureGuard generally remains within the 0.80--0.86 range. Notably, the improvements are distributed across the entire Indic language spectrum rather than being concentrated in high-resource languages such as English or Hindi. This pattern suggests that the gains stem from systematic improvements in dataset design, annotation quality, and language-specific safety coverage rather than from overfitting to a small subset of languages.

Overall, the results indicate that while CultureGuard establishes a strong multilingual baseline for culturally aware safety moderation, IndicGuard extends this foundation through deeper Indic-language specialization, resulting in improved performance across generic safety, culture-aware moderation, and adversarial jailbreak resistance.

\subsection{Quantitative Comparison}

The complete quantitative comparison between IndicGuard and CultureGuard across all languages and evaluation settings is presented in Table~\ref{tab:comparison_f1}.

\begin{table*}[t]
\centering
\small
\setlength{\tabcolsep}{4pt}
\renewcommand{\arraystretch}{0.95}
\begin{tabularx}{\textwidth}{l l|cc|cc|cc|cc|cc}
\toprule
\multirow{3}{*}{\textbf{Language}} &
\multirow{3}{*textbf{Model}} &
\multicolumn{2}{c|}{\textbf{Generic}} &
\multicolumn{2}{c|}{\textbf{CA}} &
\multicolumn{2}{c|}{\textbf{Jailbreak}} &
\multicolumn{2}{c|}{\textbf{Gen+CA}} &
\multicolumn{2}{c}{\textbf{Combined}} \\
\cmidrule(lr){3-4}\cmidrule(lr){5-6}\cmidrule(lr){7-8}\cmidrule(lr){9-10}\cmidrule(lr){11-12}
& & \scriptsize US & \scriptsize RS
  & \scriptsize US & \scriptsize RS
  & \scriptsize US & \scriptsize RS
  & \scriptsize US & \scriptsize RS
  & \scriptsize US & \scriptsize RS \\
\midrule
\multirow{2}{*}{\textbf{All Combined}}
  & \textbf{IndicGuard}  & \textbf{0.8675} & \textbf{0.8697} & \textbf{0.8510} & \textbf{0.8248} & \textbf{0.9217} & \textbf{0.9361} & \textbf{0.8642} & \textbf{0.8601} & \textbf{0.8808} & \textbf{0.8847} \\
  & \textbf{CultureGuard} & \textbf{0.8179} & \textbf{0.8522} & \textbf{0.8176} & \textbf{0.7939} & \textbf{0.8804} & \textbf{0.9025} & \textbf{0.8164} & \textbf{0.8261} & \textbf{0.8190} & \textbf{0.8261} \\
\midrule

\multirow{2}{*}{English}
  & IndicGuard  & 0.8717 & 0.8751 & 0.8471 & 0.8390 & 0.9508 & 0.9480 & 0.8668 & 0.8675 & 0.8910 & 0.8936 \\
  & CultureGuard & 0.8337 & 0.8706 & 0.8113 & 0.8259 & 0.9531 & 0.9449 & 0.8215 & 0.8507 & 0.8596 & 0.8507 \\
\midrule

\multirow{2}{*}{Bengali}
  & IndicGuard  & 0.8626 & 0.8754 & 0.8492 & 0.8179 & 0.9228 & 0.9340 & 0.8599 & 0.8631 & 0.8780 & 0.8860 \\
  & CultureGuard & 0.8128 & 0.8343 & 0.8200 & 0.7848 & 0.9022 & 0.9285 & 0.8158 & 0.8123 & 0.8247 & 0.8123 \\
\midrule

\multirow{2}{*}{Gujarati}
  & IndicGuard  & 0.8753 & 0.8593 & 0.8542 & 0.8135 & 0.9239 & 0.9420 & 0.8712 & 0.8494 & 0.8864 & 0.8794 \\
  & CultureGuard & 0.8179 & 0.8670 & 0.8166 & 0.7704 & 0.8747 & 0.9150 & 0.8164 & 0.8243 & 0.8182 & 0.8243 \\
\midrule

\multirow{2}{*}{Hindi}
  & IndicGuard  & 0.8743 & 0.8691 & 0.8487 & 0.8310 & 0.9318 & 0.9519 & 0.8692 & 0.8608 & 0.8872 & 0.8903 \\
  & CultureGuard & 0.8128 & 0.8527 & 0.8219 & 0.8084 & 0.9179 & 0.9494 & 0.8168 & 0.8331 & 0.8308 & 0.8331 \\
\midrule

\multirow{2}{*}{Kannada}
  & IndicGuard  & 0.8711 & 0.8739 & 0.8590 & 0.8303 & 0.9066 & 0.9180 & 0.8687 & 0.8646 & 0.8796 & 0.8819 \\
  & CultureGuard & 0.8159 & 0.8455 & 0.8340 & 0.7887 & 0.8658 & 0.8865 & 0.8242 & 0.8203 & 0.8157 & 0.8203 \\
\midrule

\multirow{2}{*}{Malayalam}
  & IndicGuard  & 0.8589 & 0.8696 & 0.8463 & 0.8355 & 0.9159 & 0.9297 & 0.8565 & 0.8622 & 0.8736 & 0.8840 \\
  & CultureGuard & 0.8130 & 0.8280 & 0.8080 & 0.7992 & 0.8237 & 0.8553 & 0.8102 & 0.8151 & 0.7891 & 0.8151 \\
\midrule

\multirow{2}{*}{Marathi}
  & IndicGuard  & 0.8706 & 0.8692 & 0.8625 & 0.8399 & 0.9280 & 0.9359 & 0.8691 & 0.8629 & 0.8860 & 0.8865 \\
  & CultureGuard & 0.8158 & 0.8596 & 0.8266 & 0.8079 & 0.8948 & 0.9292 & 0.8205 & 0.8367 & 0.8277 & 0.8367 \\
\midrule

\multirow{2}{*}{Punjabi}
  & IndicGuard  & 0.8640 & 0.8715 & 0.8505 & 0.8132 & 0.9269 & 0.9360 & 0.8614 & 0.8590 & 0.8802 & 0.8839 \\
  & CultureGuard & 0.8128 & 0.8310 & 0.8127 & 0.8012 & 0.8438 & 0.8650 & 0.8123 & 0.8179 & 0.8046 & 0.8179 \\
\midrule

\multirow{2}{*}{Tamil}
  & IndicGuard  & 0.8642 & 0.8657 & 0.8442 & 0.8262 & 0.9029 & 0.9320 & 0.8601 & 0.8573 & 0.8724 & 0.8815 \\
  & CultureGuard & 0.8227 & 0.8516 & 0.8213 & 0.7947 & 0.8631 & 0.9144 & 0.8214 & 0.8266 & 0.8184 & 0.8266 \\
\midrule

\multirow{2}{*}{Telugu}
  & IndicGuard  & 0.8690 & 0.8710 & 0.8455 & 0.8051 & 0.9207 & 0.9319 & 0.8643 & 0.8571 & 0.8806 & 0.8814 \\
  & CultureGuard & 0.8098 & 0.8569 & 0.8024 & 0.7632 & 0.8604 & 0.8906 & 0.8051 & 0.8155 & 0.8132 & 0.8155 \\
\midrule

\multirow{2}{*}{Urdu}
  & IndicGuard  & 0.8605 & 0.8666 & 0.8542 & 0.8217 & 0.9089 & 0.9380 & 0.8595 & 0.8570 & 0.8737 & 0.8833 \\
  & CultureGuard & 0.8233 & 0.8530 & 0.8085 & 0.7992 & 0.8545 & 0.8495 & 0.8157 & 0.8293 & 0.8016 & 0.8293 \\
\bottomrule
\end{tabularx}
\vspace{\abovecaptionskip}
\caption{\textbf{IndicGuard vs.\ CultureGuard --- Macro $F_1$:} User Safety (US) and Response Safety (RS) F1 scores across all evaluation settings.}
\label{tab:comparison_f1}
\vspace{\belowcaptionskip}
\end{table*}

\section{Detailed Performance Metrics}
\label{sec:appendix_tables}
This section reports the full performance metrics for all 11 languages across the six configurations, along with aggregate summaries and delta improvements evaluated in this study.

\subsection{Interpretation of Empirical Values}
The empirical results across the evaluated setups demonstrate strong, well-balanced multilingual safety moderation performance across all 11 languages. Under the fully \textit{Combined} distribution, the multilingual model achieves a robust mean \textit{User Safety} $F_1$ score of 0.8800 and a mean \textit{Response Safety} $F_1$ score of 0.8846. English achieves the highest overall scores (0.8910 for User and 0.8936 for Response Safety), while the remaining ten Indic languages cluster closely behind with limited variance. Specifically, when excluding English, the non-English mean $F_1$ score is 0.8789 for User Safety and 0.8837 for Response Safety. This translates to a modest English–Indic performance gap of merely +0.0121 for user-side prompts and +0.0099 for model responses. These thin margins demonstrate that the training pipeline successfully mitigates the cross-lingual performance degradation typically found in English-centric architectures. Furthermore, the marginal performance disparity between input and output classifications ($0.0046$) indicates a high degree of turn-invariant moderation consistency and highlights a finely calibrated decision threshold.  \subsection{Supporting Ablation Analysis}
The accompanying ablation study systematically quantifies the marginal contribution of each structural data component—\textit{Generic}, \textit{Culture-Adaptive (CA)}, and \textit{Jailbreaking}—by incrementally expanding the training configuration. When evaluating performance regimes in isolation, adversarial \textit{Jailbreak} robustness emerges as the strongest, yielding an isolated mean $F_1$ score of 0.9225 for User Safety and 0.9360 for Response Safety. Conversely, the \textit{Culture-Adaptive} split proves to be the most challenging domain, returning a lower standalone mean $F_1$ of 0.8516 (User) and 0.8246 (Response) compared to the \textit{Generic} baselines (0.8673 and 0.8691, respectively). This drop highlights that uncovering subtle, localized, and culturally nuanced socio-political harms demands much finer semantic discrimination than detecting explicit policy violations.  \subsection{Data Composition and Generalization Effects}
Crucially, the ablation highlights how incorporating these distinct data layers affects joint performance without causing catastrophic forgetting. Mixing generic safety data with localized examples in the \textit{Gen+CA} configuration yields performance metrics (0.8651 User, 0.8604 Response) that track closely alongside the baseline \textit{Generic} model. This confirms that exposing the model to regional socio-cultural concepts preserves general policy alignment. Ultimately, moving from isolated data subsets to the fully unified \textit{Combined} multi-regime training unlocks notable absolute improvements of +0.0284 on the user side and +0.0600 on the response side over the \textit{CA} setup alone. This sharp recovery, particularly on response safety classification, indicates that unified training promotes shared cross-lingual representations that effectively bridge and generalize across heterogeneous safety phenomena.

\setlength{\textfloatsep}{10pt plus 2pt minus 4pt}
\setlength{\floatsep}{8pt plus 2pt minus 2pt}
\setlength{\intextsep}{8pt plus 2pt minus 2pt}
\setlength{\abovecaptionskip}{5pt}
\setlength{\belowcaptionskip}{2pt}

\begin{table*}[t]
\centering
\small
\setlength{\tabcolsep}{5pt}
\begin{tabularx}{\textwidth}{l|ccc|ccc}
\toprule
\multirow{2}{*}{\textbf{Evaluation Setting}} & \multicolumn{3}{c|}{\textbf{Mean Baseline Performance (F1)}} & \multicolumn{3}{c}{\textbf{Marginal Improvement Delta ($\Delta$)}} \\
\cmidrule(lr){2-4} \cmidrule(lr){5-7}
& User Safety & Response Safety & Combined Mean & User Delta & Response Delta & Total $\Delta$ \\
\midrule
Generic Only (Baseline) & 0.8673 & 0.8691 & 0.8682 & --- & --- & --- \\
Culture-Adaptive (Gen+CA) & 0.8651 & 0.8604 & 0.8628 & $-0.0022$ & $-0.0087$ & $-0.0054$ \\
Full Corpus (Gen+CA+JB)   & 0.8800 & 0.8846 & 0.8823 & $+0.0149$ & $+0.0242$ & $+0.0195$ \\
\bottomrule
\end{tabularx}
\vspace{\abovecaptionskip}
\caption{Aggregate Performance Summary and Marginal Improvement Deltas Across Evaluation Settings.}
\label{tab:merged_summary_deltas}
\vspace{\belowcaptionskip}
\end{table*}

\begin{table*}[t]
\centering
\small
\setlength{\tabcolsep}{4pt}
\renewcommand{\arraystretch}{0.95}
\begin{tabularx}{\textwidth}{l|cccc|cccc}
\toprule
\multirow{2}{*}{\textbf{Language}} &
\multicolumn{4}{c|}{\textbf{User Safety}} &
\multicolumn{4}{c}{\textbf{Response Safety}} \\
\cmidrule(lr){2-5}\cmidrule(lr){6-9}
& Acc. & Prec. & Rec. & F1
& Acc. & Prec. & Rec. & F1 \\
\midrule
Combined All Languages & 0.8188 & 0.8228 & 0.8188 & 0.8204 & 0.7852 & 0.7849 & 0.7852 & 0.7849 \\
English   & 0.8798 & 0.8808 & 0.8798 & 0.8800 & 0.8889 & 0.9013 & 0.8889 & 0.8399 \\
Gujarati  & 0.8518 & 0.8519 & 0.8518 & 0.8519 & 0.8327 & 0.8336 & 0.8327 & 0.8328 \\
Hindi     & 0.8466 & 0.8521 & 0.8466 & 0.8449 & 0.8214 & 0.8330 & 0.8214 & 0.8205 \\
Kannada   & 0.8444 & 0.8484 & 0.8444 & 0.8424 & 0.7992 & 0.8218 & 0.7992 & 0.7937 \\
Malayalam & 0.8533 & 0.8564 & 0.8533 & 0.8536 & 0.8602 & 0.8737 & 0.8602 & 0.8625 \\
Marathi   & 0.8610 & 0.8610 & 0.8610 & 0.8610 & 0.8635 & 0.8640 & 0.8635 & 0.8635 \\
Urdu      & 0.8600 & 0.8611 & 0.8600 & 0.8601 & 0.8536 & 0.8536 & 0.8536 & 0.8536 \\
Tamil     & 0.8625 & 0.8624 & 0.8625 & 0.8624 & 0.8473 & 0.8490 & 0.8473 & 0.8473 \\
Punjabi   & 0.8585 & 0.8587 & 0.8585 & 0.8585 & 0.8512 & 0.8539 & 0.8512 & 0.8511 \\
Telugu    & 0.8676 & 0.8682 & 0.8676 & 0.8677 & 0.8561 & 0.8562 & 0.8561 & 0.8561 \\
Bengali   & 0.8605 & 0.8610 & 0.8605 & 0.8606 & 0.8622 & 0.8623 & 0.8622 & 0.8623 \\
\bottomrule
\end{tabularx}
\caption{\textbf{Generic Model (Combined):} Trained on combined generic data of all 11 languages; tested on the aggregate dataset.}
\label{tab:combined_gen}
\end{table*}

\begin{table*}[t]
\centering
\small
\setlength{\tabcolsep}{4pt}
\renewcommand{\arraystretch}{0.95}
\begin{tabularx}{\textwidth}{l|cccc|cccc}
\toprule
\multirow{2}{*}{\textbf{Language}} &
\multicolumn{4}{c|}{\textbf{User Safety}} &
\multicolumn{4}{c}{\textbf{Response Safety}} \\
\cmidrule(lr){2-5}\cmidrule(lr){6-9}
& Acc. & Prec. & Rec. & F1
& Acc. & Prec. & Rec. & F1 \\
\midrule
Combined All Languages & 0.8196 & 0.8234 & 0.8196 & 0.8212 & 0.7843 & 0.7839 & 0.7843 & 0.7840 \\
English   & 0.8733 & 0.8737 & 0.8733 & 0.8734 & 0.8710 & 0.8877 & 0.8710 & 0.8135 \\
Gujarati  & 0.8485 & 0.8489 & 0.8485 & 0.8487 & 0.8256 & 0.8261 & 0.8256 & 0.8257 \\
Hindi     & 0.8435 & 0.8477 & 0.8435 & 0.8411 & 0.8127 & 0.8247 & 0.8127 & 0.8119 \\
Kannada   & 0.8413 & 0.8426 & 0.8413 & 0.8389 & 0.7926 & 0.8124 & 0.7926 & 0.7877 \\
Malayalam & 0.8529 & 0.8548 & 0.8529 & 0.8527 & 0.8499 & 0.8628 & 0.8499 & 0.8523 \\
Marathi   & 0.8566 & 0.8566 & 0.8566 & 0.8566 & 0.8516 & 0.8520 & 0.8516 & 0.8517 \\
Urdu      & 0.8530 & 0.8548 & 0.8530 & 0.8534 & 0.8430 & 0.8431 & 0.8430 & 0.8428 \\
Tamil     & 0.8558 & 0.8556 & 0.8558 & 0.8556 & 0.8351 & 0.8367 & 0.8351 & 0.8352 \\
Punjabi   & 0.8550 & 0.8552 & 0.8550 & 0.8551 & 0.8410 & 0.8429 & 0.8410 & 0.8411 \\
Telugu    & 0.8603 & 0.8614 & 0.8603 & 0.8606 & 0.8391 & 0.8391 & 0.8391 & 0.8391 \\
Bengali   & 0.8574 & 0.8584 & 0.8574 & 0.8577 & 0.8478 & 0.8478 & 0.8478 & 0.8478 \\
\bottomrule
\end{tabularx}
\caption{\textbf{Gen+CA Model (Combined):} Trained on combined generic and culture-adaptive data of all 11 languages.}
\label{tab:combined_gen_ca}
\end{table*}

\begin{table*}[t]
\centering
\small
\setlength{\tabcolsep}{4pt}
\renewcommand{\arraystretch}{0.95}
\begin{tabularx}{\textwidth}{l|cccc|cccc}
\toprule
\multirow{2}{*}{\textbf{Language}} &
\multicolumn{4}{c|}{\textbf{User Safety}} &
\multicolumn{4}{c}{\textbf{Response Safety}} \\
\cmidrule(lr){2-5}\cmidrule(lr){6-9}
& Acc. & Prec. & Rec. & F1
& Acc. & Prec. & Rec. & F1 \\
\midrule
\textbf{All Combined Avg.} 
& \textbf{0.8469} & \textbf{0.8486} & \textbf{0.8469} & \textbf{0.8461}
& \textbf{0.8402} & \textbf{0.8458} & \textbf{0.8402} & \textbf{0.8344} \\
English   & 0.8607 & 0.8644 & 0.8607 & 0.8606 & 0.8759 & 0.8914 & 0.8759 & 0.8204 \\
Gujarati  & 0.8420 & 0.8420 & 0.8420 & 0.8420 & 0.8321 & 0.8327 & 0.8321 & 0.8322 \\
Hindi     & 0.8451 & 0.8514 & 0.8451 & 0.8418 & 0.8271 & 0.8408 & 0.8271 & 0.8260 \\
Kannada   & 0.8309 & 0.8352 & 0.8309 & 0.8262 & 0.7900 & 0.8127 & 0.7900 & 0.7822 \\
Malayalam & 0.8554 & 0.8576 & 0.8554 & 0.8548 & 0.8373 & 0.8394 & 0.8373 & 0.8379 \\
Marathi   & 0.8489 & 0.8489 & 0.8489 & 0.8489 & 0.8607 & 0.8610 & 0.8607 & 0.8607 \\
Urdu      & 0.8423 & 0.8444 & 0.8423 & 0.8428 & 0.8444 & 0.8451 & 0.8444 & 0.8442 \\
Tamil     & 0.8466 & 0.8463 & 0.8466 & 0.8459 & 0.8420 & 0.8454 & 0.8420 & 0.8419 \\
Punjabi   & 0.8426 & 0.8422 & 0.8426 & 0.8423 & 0.8429 & 0.8450 & 0.8429 & 0.8429 \\
Telugu    & 0.8504 & 0.8508 & 0.8504 & 0.8506 & 0.8371 & 0.8372 & 0.8371 & 0.8371 \\
Bengali   & 0.8507 & 0.8511 & 0.8507 & 0.8508 & 0.8531 & 0.8531 & 0.8531 & 0.8531 \\
\bottomrule
\end{tabularx}
\caption{\textbf{Gen+CA+JB Model (Combined):} Trained on combined generic, culture-adaptive, and jailbreaking data of all 11 languages.}
\label{tab:combined_all}
\end{table*}

\begin{table*}[t]
\centering
\small
\setlength{\tabcolsep}{4pt}
\renewcommand{\arraystretch}{0.95}
\begin{tabularx}{\textwidth}{l|cccc|cccc}
\toprule
\multirow{2}{*}{\textbf{Language}} &
\multicolumn{4}{c|}{\textbf{User Safety}} &
\multicolumn{4}{c}{\textbf{Response Safety}} \\
\cmidrule(lr){2-5}\cmidrule(lr){6-9}
& Acc. & Prec. & Rec. & F1
& Acc. & Prec. & Rec. & F1 \\
\midrule
Combined All Languages & 0.8302 & 0.8352 & 0.8302 & 0.8317 & 0.8189 & 0.8192 & 0.8189 & 0.8189 \\
English   & 0.8717 & 0.8719 & 0.8717 & 0.8714 & 0.8573 & 0.8580 & 0.8573 & 0.8573 \\
Bengali   & 0.8518 & 0.8526 & 0.8518 & 0.8513 & 0.8438 & 0.8494 & 0.8438 & 0.8435 \\
Gujarati  & 0.8498 & 0.8509 & 0.8498 & 0.8491 & 0.8098 & 0.8133 & 0.8098 & 0.8097 \\
Hindi     & 0.8396 & 0.8396 & 0.8396 & 0.8393 & 0.8354 & 0.8382 & 0.8354 & 0.8353 \\
Kannada   & 0.8488 & 0.8494 & 0.8488 & 0.8483 & 0.8221 & 0.8256 & 0.8221 & 0.8220 \\
Malayalam & 0.8462 & 0.8469 & 0.8462 & 0.8457 & 0.8280 & 0.8337 & 0.8280 & 0.8277 \\
Marathi   & 0.8529 & 0.8539 & 0.8529 & 0.8522 & 0.8174 & 0.8216 & 0.8174 & 0.8172 \\
Punjabi   & 0.8488 & 0.8492 & 0.8488 & 0.8483 & 0.8258 & 0.8302 & 0.8258 & 0.8256 \\
Tamil     & 0.8447 & 0.8448 & 0.8447 & 0.8444 & 0.8258 & 0.8282 & 0.8258 & 0.8257 \\
Telugu    & 0.8544 & 0.8561 & 0.8544 & 0.8536 & 0.8380 & 0.8415 & 0.8380 & 0.8379 \\
Urdu      & 0.8427 & 0.8435 & 0.8427 & 0.8420 & 0.8303 & 0.8364 & 0.8303 & 0.8299 \\
\bottomrule
\end{tabularx}
\caption{\textbf{Generic Models (Individual):} Each model trained on a single language using only generic data and evaluated on that language's test set.}
\label{tab:generic_lang}
\end{table*}

\begin{table*}[t]
\centering
\small
\setlength{\tabcolsep}{4pt}
\renewcommand{\arraystretch}{0.95}
\begin{tabularx}{\textwidth}{l|cccc|cccc}
\toprule
\multirow{2}{*}{\textbf{Language}} &
\multicolumn{4}{c|}{\textbf{User Safety}} &
\multicolumn{4}{c}{\textbf{Response Safety}} \\
\cmidrule(lr){2-5}\cmidrule(lr){6-9}
& Acc. & Prec. & Rec. & F1
& Acc. & Prec. & Rec. & F1 \\
\midrule
English   & 0.8578 & 0.8577 & 0.8578 & 0.8577 & 0.8401 & 0.8402 & 0.8401 & 0.8401 \\
Bengali   & 0.8469 & 0.8466 & 0.8469 & 0.8463 & 0.8353 & 0.8395 & 0.8353 & 0.8352 \\
Gujarati  & 0.8457 & 0.8457 & 0.8457 & 0.8448 & 0.7971 & 0.7991 & 0.7971 & 0.7972 \\
Hindi     & 0.8360 & 0.8356 & 0.8360 & 0.8357 & 0.8239 & 0.8253 & 0.8239 & 0.8240 \\
Kannada   & 0.8445 & 0.8441 & 0.8445 & 0.8440 & 0.8106 & 0.8126 & 0.8106 & 0.8107 \\
Malayalam & 0.8392 & 0.8388 & 0.8392 & 0.8387 & 0.8162 & 0.8200 & 0.8162 & 0.8161 \\
Marathi   & 0.8477 & 0.8477 & 0.8477 & 0.8469 & 0.8079 & 0.8111 & 0.8079 & 0.8079 \\
Punjabi   & 0.8420 & 0.8417 & 0.8420 & 0.8415 & 0.8183 & 0.8214 & 0.8183 & 0.8183 \\
Tamil     & 0.8352 & 0.8348 & 0.8352 & 0.8349 & 0.8192 & 0.8211 & 0.8192 & 0.8193 \\
Telugu    & 0.8481 & 0.8481 & 0.8481 & 0.8473 & 0.8250 & 0.8271 & 0.8250 & 0.8251 \\
Urdu      & 0.8384 & 0.8381 & 0.8384 & 0.8377 & 0.8208 & 0.8246 & 0.8208 & 0.8208 \\
\bottomrule
\end{tabularx}
\caption{\textbf{Gen+CA Models (Individual):} Each model trained per language using combined generic and culture-adaptive data.}
\label{tab:gen_ca_lang}
\end{table*}

\begin{table*}[t]
\centering
\small
\setlength{\tabcolsep}{4pt}
\renewcommand{\arraystretch}{0.95}
\begin{tabularx}{\textwidth}{l|cccc|cccc}
\toprule
\multirow{2}{*}{\textbf{Language}} &
\multicolumn{4}{c|}{\textbf{User Safety}} &
\multicolumn{4}{c}{\textbf{Response Safety}} \\
\cmidrule(lr){2-5}\cmidrule(lr){6-9}
& Acc. & Prec. & Rec. & F1
& Acc. & Prec. & Rec. & F1 \\
\midrule
English   & 0.8636 & 0.8645 & 0.8636 & 0.8639 & 0.8491 & 0.8491 & 0.8491 & 0.8491 \\
Bengali   & 0.8455 & 0.8452 & 0.8455 & 0.8453 & 0.8414 & 0.8440 & 0.8414 & 0.8413 \\
Gujarati  & 0.8455 & 0.8451 & 0.8455 & 0.8450 & 0.8122 & 0.8148 & 0.8122 & 0.8121 \\
Hindi     & 0.8420 & 0.8423 & 0.8420 & 0.8421 & 0.8408 & 0.8418 & 0.8408 & 0.8408 \\
Kannada   & 0.8412 & 0.8412 & 0.8412 & 0.8412 & 0.8207 & 0.8224 & 0.8207 & 0.8207 \\
Malayalam & 0.8423 & 0.8422 & 0.8423 & 0.8423 & 0.8285 & 0.8322 & 0.8285 & 0.8283 \\
Marathi   & 0.8510 & 0.8506 & 0.8510 & 0.8507 & 0.8209 & 0.8234 & 0.8209 & 0.8208 \\
Punjabi   & 0.8420 & 0.8418 & 0.8420 & 0.8419 & 0.8240 & 0.8280 & 0.8240 & 0.8238 \\
Tamil     & 0.8360 & 0.8362 & 0.8360 & 0.8361 & 0.8266 & 0.8284 & 0.8266 & 0.8266 \\
Telugu    & 0.8449 & 0.8445 & 0.8449 & 0.8445 & 0.8279 & 0.8307 & 0.8279 & 0.8278 \\
Urdu      & 0.8366 & 0.8364 & 0.8366 & 0.8364 & 0.8322 & 0.8344 & 0.8322 & 0.8322 \\
\bottomrule
\end{tabularx}
\caption{\textbf{Gen+CA+JB Models (Individual):} Each model trained per language using all data types: generic, culture-adaptive, and jailbreaking.}
\label{tab:gen_ca_jb_lang}
\end{table*}

\end{document}